\documentclass{article}
\usepackage{ijcai20}

\usepackage{times}
\usepackage{soul}
\usepackage{url}
\usepackage[hidelinks]{hyperref}
\usepackage[utf8]{inputenc}
\usepackage[small]{caption}
\usepackage{graphicx}
\usepackage{amsmath}
\usepackage{amssymb}
\usepackage{amsthm}
\usepackage{amsfonts}
\usepackage{booktabs}
\usepackage{algorithm}
\usepackage{algorithmic}
\usepackage{multirow}

\title{Exploiting Visual Semantic Reasoning for Video-Text Retrieval}

\author{
	Zerun Feng$^1$\and
	Zhimin Zeng$^{1,2}$\and
	Caili Guo$^{1,2}$\footnote{Corresponding author}\And
	Zheng Li$^1$
	\affiliations
	$^1$Beijing Key Laboratory of Network System Architecture and Convergence,\\
	School of Information and Communication Engineering,\\
	Beijing University of Posts and Telecommunications, Beijing, China\\
	$^2$Beijing Laboratory of Advanced Information Networks, Beijing, China\\
	\emails
	\{fengzerun,zengzm, guocaili, lizhengzachary\}@bupt.edu.cn
}

\begin{document}
	
	\maketitle
	
	\begin{abstract}
		Video retrieval is a challenging research topic bridging the vision and language areas and has attracted broad attention in recent years. Previous works have been devoted to representing videos by directly encoding from frame-level features. In fact, videos consist of various and abundant semantic relations to which existing methods pay less attention. To address this issue, we propose a Visual Semantic Enhanced Reasoning Network (ViSERN) to exploit reasoning between frame regions. Specifically, we consider frame regions as vertices and construct a fully-connected semantic correlation graph. Then, we perform reasoning by novel random walk rule-based graph convolutional networks to generate region features involved with semantic relations. With the benefit of reasoning, semantic interactions between regions are considered, while the impact of redundancy is suppressed. Finally, the region features are aggregated to form frame-level features for further encoding to measure video-text similarity. Extensive experiments on two public benchmark datasets validate the effectiveness of our method by achieving state-of-the-art performance due to the powerful semantic reasoning.
	\end{abstract}
	
	\section{Introduction}
	Vision and language are two fundamental bridges for humans to access the real world. In the research community, video retrieval is one of the active topics to make connections between these two counterparts, which aims to measure the semantic similarity between a video and its corresponding caption in an embedding space. Semantic-based video retrieval can reduce the effort of manual annotations to improve productiveness.
	
	\begin{figure}[t]
		\centering
		\includegraphics[width=1.0\linewidth]{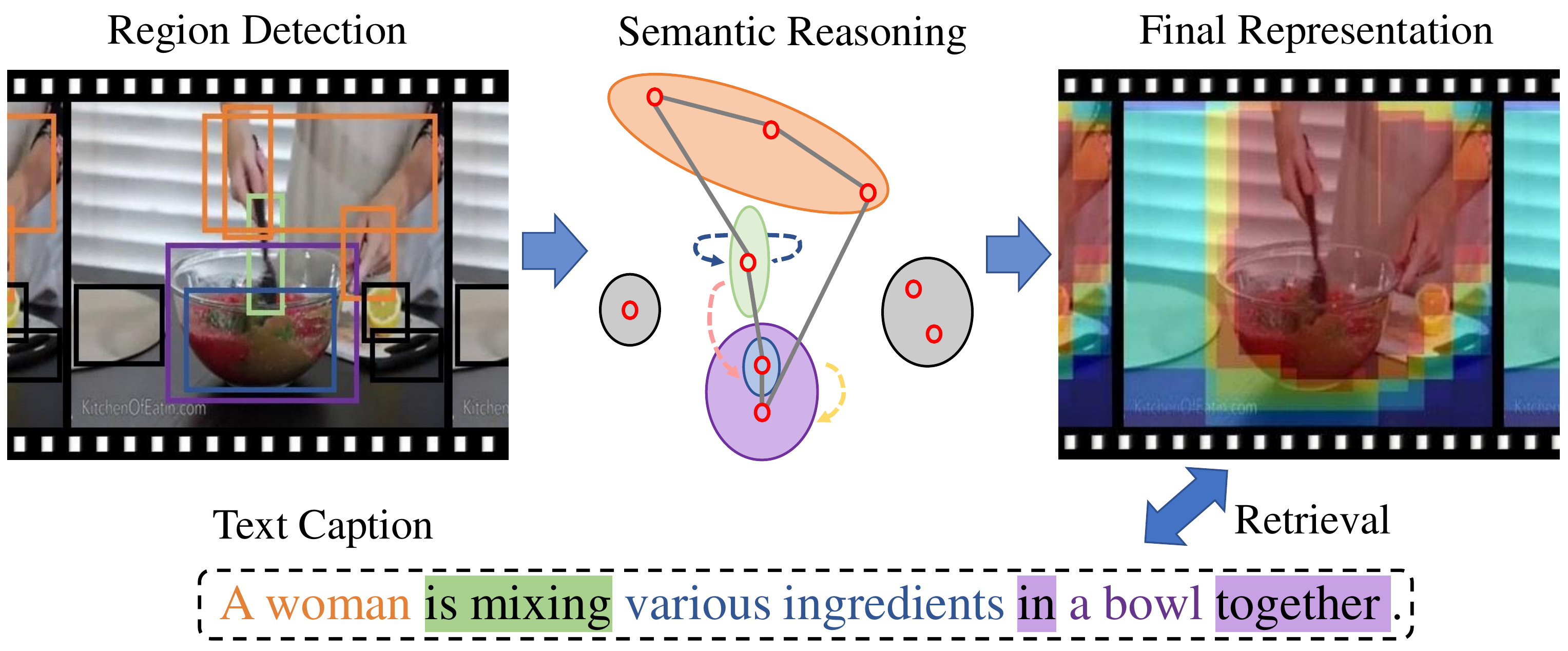}
		\caption{Illustration of the proposed Visual Semantic Enhanced Reasoning Network (ViSERN).}    
		\label{fig:schematic-diagram}
	\end{figure}
	
	Lately, with the dramatic development of visual understanding tasks, various video retrieval methods have been proposed and achieved promising performance. Most of them~\cite{mithun2018learning,dong2018predicting} utilize pre-trained Convolutional Neural Network (CNN) to extract mid-level features as the representations of frames for further manipulating to calculate similarity scores between videos and texts. More recently,~\cite{dong2019dual} encode frame-level features in a multi-level way for video retrieval. Because of the existing huge video-text representation discrepancy, it is still an enormous challenge to predict the accurate video-text retrieval results. To lift this limitation in a principled way, we need to take a deeper insight into videos. The most basic unit of videos is a series of regions (such as persons, objects and landscapes). We human beings understand the videos by observing and reasoning semantic relation evolution between these regions during the timeline.  Recall that if we describe video contents using a sentence, we will not only focus on objects, but also care about motions, interactions, relative positions and other high-level semantic concepts (e.g., ``mixing", ``in" and ``together", the highlighted words shown in Figure~\ref{fig:schematic-diagram}), which lack apparent correspondence between visual representations and word descriptions. Moreover, videos have spatial-temporal redundancy information (e.g., ``lemon" and ``chopping board" denoted by black boxes shown in Figure~\ref{fig:schematic-diagram}), which is not always mentioned in captions. The intuitionistic phenomena indicate that visual semantic reasoning is fundamental for humans to understand the real world~\cite{katsuki2014bottom}. However, the current existing video retrieval methods lack such kind of reasoning mechanism. Although these representations can obtain global visual information from the frames, they fail to capture high-level semantic interactions and treat discriminative regions equally. Thus, the semantic gap between videos and texts is not well addressed,  and it leads to limited performance when matching videos and texts.
	
	In this paper, we propose a Visual Semantic Enhanced Reasoning Network (ViSERN) to generate visual representations by exploiting semantic relations. After sampling frames, our model detects frame regions by bottom-up attention~\cite{anderson2018bottom} and extract region features. In this way, each frame can be represented by several regions. Specifically, the bottom-up attention module is implemented with Faster R-CNN~\cite{ren2015faster} pre-trained on Visual Genome~\cite{krishna2017visual}, an image region relation annotated dataset. Then, we build up a fully-connected graph with these regions as vertices and their semantic relevance as edges. We perform reasoning by novel random walk~\cite{perozzi2014deepwalk} rule-based Graph Convolutional Networks (GCN)~\cite{kipf2016semi} to produce region features with semantic relations.
	
	We suppose that the regions with intensely semantic relations would be beneficial to infer video-text similarity. Therefore, we combine random walks with GCN and perform reasoning to enhance the effect of the most valuable ones. The random walks are a rule for vertices to access their neighbors. The transition probability is determined by the weights of edges. The neighbor vertices with higher weight are more likely to be visited. By this means, we can choose the discriminative regions and drop the insignificant ones by performing reasoning according to the random walk rule. Finally, the region features are passed by mean pooling layer to generate frame-level features for further manipulating to obtain retrieval results. 
	
	To illustrate intuitively, we also design a visualization map to analyze what has been learned after reasoning, in addition to the quantitative assessment of our method on standard benchmarks. Correlations between the frame-level feature and each reasoning region feature belonging to the same frames are calculated and visualized in an attention form. As shown in Figure~\ref{fig:schematic-diagram}, we find that pivotal semantic interactions can be well captured, while redundant frame regions cause low responses.
	
	The main contributions of our work are summarized as follows:
	
	\begin{itemize}
		
		\item We propose a novel Visual Semantic Enhanced Reasoning Network (ViSERN) based on GCN to explore semantic correlations between regions within frames for measuring video-text similarity. As far as we know, ViSERN is the first video-text retrieval model which focuses on visual semantic reasoning. 
		\item To further enhance the reasoning capacity, we observe that it is powerful to introduce the random walk rule into GCN. The regions with high semantic relevance are more possibly to access each other for better reasoning by preserving the structure graph topology.
		\item Experimental results conducted on two publicly available datasets demonstrate the effectiveness of the proposed model by achieving a new state-of-the-art due to the powerful semantic reasoning.
		
	\end{itemize}

	\section{Related Work}
	\noindent \textbf{Video Retrieval.}\quad
	Over the past several years, with big advances of deep learning in computer vision and natural language processing, we observe that embedding based methods have been proposed for video retrieval.~\cite{dong2018predicting} leverage pre-trained CNN models to extract features from frames and aggregate the frame-level features into a video-level feature by mean pooling.~\cite{dong2019dual} use mean pooling, bidirectional GRU, and 1-D CNN to encode global, temporal and local pattern information respectively for both video and text ends. However, to the best of our knowledge, no study has attempted to incorporate visual semantic reasoning when learning embedding for video retrieval. \\
	
	\noindent \textbf{Visual Semantic Reasoning.}\quad
	As an emerging research area, graph-based methods have been popular in recent years and shown to be an efficient approach to semantic relation reasoning. Graph Convolution Network (GCN)~\cite{kipf2016semi} is first proposed for semi-supervised classification. Graph convolution extends the applicable scope of standard convolution from regular grids to more general pre-defined graphs by manipulating in the spectral domain, which is also proved to be powerful in video understanding.~\cite{wang2018videos} represent videos as similarity graphs and bi-directional spatial-temporal graphs to perform reasoning via GCN for video classification.~\cite{liu2019social} design a graph model to capture social relations from videos and perform temporal reasoning with multi-scale receptive fields on the graph by GCN. Inspired by the above studies, we propose to represent the frames as fully-connected semantic correlation graphs, on which reasoning is performed by novel random walk rule-based graph convolutional networks for video retrieval. \\
	
	\noindent \textbf{Random Walks.}\quad
	Random walks on a graph is a rule to visit a sequence of vertices together with a sequence of edges and widely applied in network representation learning.~\cite{perozzi2014deepwalk} treat nodes as words and sentences by truncated random walks to bridge the gap between network embeddings and word embeddings.~\cite{grover2016node2vec} define a flexible notion of a node's network neighborhood and design a biased random walk procedure for learning representations in complex networks. Since networks and graphs have similar topological structures, we argue that it is reasonable to introduce random walk statistics into GCN to enhance its visual reasoning ability. \\
	
	\begin{figure*}
		\centering
		\includegraphics[width=1.0\linewidth, , height=0.237\textheight]{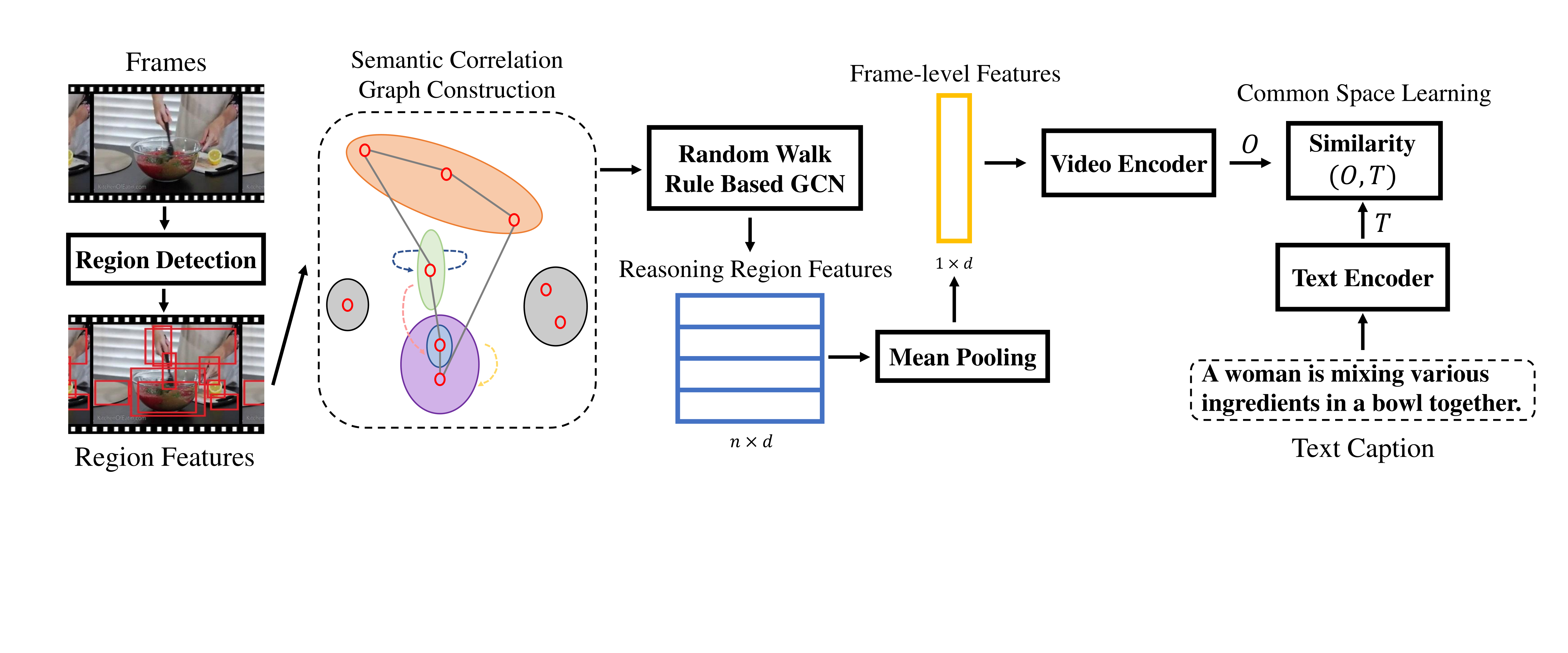}
		\caption{Architecture of the proposed Visual Semantic Enhanced Reasoning Network.}
		\label{fig:framework}
	\end{figure*}
   	
	\section{Proposed Method}
	In order to capture visual semantic correlations for video feature representations, we propose a Visual Semantic Enhanced Reasoning Network (ViSERN). The architecture of ViSERN is shown in Figure~\ref{fig:framework}. Concretely, our goal is to measure the similarity between video-text pairs by common space learning. We utilize the bottom-up attention model to generate frame regions and extract features from frames (Sec.~\ref{Video Key Frames Representation}). For the regions, we construct a graph to model semantic correlations (Sec.~\ref{Semantic Relation Graph Construction}). Subsequently, we do semantic reasoning between these regions by leveraging random walk rule-based Graph Convolutional Networks (GCN) to generate region features with relation information (Sec.~\ref{Reasoning by Random Walk Rule Based GCN}). Finally, video and text features are generated, and the whole model is trained with common space learning (Sec.~\ref{Common Space Learning}).
	
	\subsection{Region Features Extraction}\label{Video Key Frames Representation}
	After sampling frames from videos, we use the bottom-up attention model~\cite{anderson2018bottom} to extract a set of salient frame regions (e.g., person, cars and sky). We implement the bottom-up attention model with the same configuration in~\cite{lee2018stacked}. The model is pre-trained on Visual Genomes~\cite{krishna2017visual}, a large-scale image dataset annotating the interactions/relationships between objects. Objects and their relationships within each image in the dataset are densely annotated. Specifically, with the benefit of the bottom-up attention model, each frame can be represented by a set of features with semantic implications. The top $n$ regions with the highest detection confidence scores are selected. We extract each region feature with $d$ dimensions after \textit{pool5} layer of ResNet-101~\cite{he2016deep} as backbone in the model.
	
	In this way, we represent each frame by using its region features $V = \{v_1,...,v_n\}, v_i \in \mathbb{R}^d$.
	
	\subsection{Semantic Correlation Graph Construction}\label{Semantic Relation Graph Construction}
	In our work, a relation graph is designed to conduct semantic reasoning between regions. Particularly, we construct a fully-connected semantic correlation graph $G = (V, E)$ on a frame with $n$ region features.
	
	In the graph, $V = \{v_1,...,v_n\}, v_i \in \mathbb{R}^d$ is the collection of region features as depicted above. As for edge set $E$, each two vertices exist a link, we suppose $\forall (v_i, v_j) \in E $. To construct an adjacency matrix $R$ representing the edge set $E$, dot-product attention mechanism~\cite{vaswani2017attention} is implemented in an embedding space. We formulate $R$ in a form of symmetric adjacency matrix following the definition of undirected graph to measure semantic relations between regions:
	
	\begin{equation}\label{adjacency matrix}
	R(v_i, v_j) = \phi(v_i)^T\theta(v_j) + \theta(v_i)^T\phi(v_j)\,,
	\end{equation}
	where $\phi(\cdot) = W_{\phi}(\cdot) + b_{\phi}$ and $\theta(\cdot) = W_{\theta}(\cdot) + b_{\theta}$ are two linear embeddings with bias. In this way, the two regions can be connected with a higher weight edge if they have more intensely semantic correlations. 
	
	\subsection{Reasoning by Random Walk Rule-Based GCN}\label{Reasoning by Random Walk Rule Based GCN}
	At the heart of our approach, we propose a module based on Graph Convolutional Networks (GCN)~\cite{kipf2016semi} to perform semantic reasoning between regions. To process region features in a spectral domain, we consider spectral convolutions on the graph $G$. For concise descriptions, we start with the set of vertices $x \in \mathbb{R}^n$, i.e., a scalar for all $n$ vertices, then generalize to multi-input channels.
	
	We recall that a fundamental operator in spectral graph analysis is the graph Laplacian. The combinatorial definition of graph Laplacian matrix is $L = D - R \in \mathbb{R}^{n \times n}$ where $D \in \mathbb{R}^{n \times n}$ is diagonal degree matrix with $D_{ii} = \sum\nolimits_j R_{ij}$. Moreover, the symmetric normalized graph Laplacian matrix is $L^{sym} = I_n - D^{-1/2}RD^{-1/2}$, where $I_n \in \mathbb{R}^{n \times n}$ is the identity matrix. The convolution operator on vertices $\star x$ are filtered by $g_{\theta} = diag(\theta)$~\cite{bruna2014spectral},  parameterized by learnable $\theta \in \mathbb{R}^n$ in the spectral domain:
	
	\begin{equation}
	g_{\theta} \star x = Ug_{\theta}U^Tx\,,
	\end{equation}
	where $L^{sym} = U\Lambda U^T$ is diagonalized by the spectral basis $U \in \mathbb{R}^{n \times n}$, with a diagonal matrix of its eigenvalues $\Lambda \in \mathbb{R}^{n \times n}$. We can understand this convolution operation by three steps: First, $x$ is transformed from the spatial domain into the spectral domain by $U^Tx$. Next, the computation in the spectral domain is evaluated through $g_{\theta}U^Tx$. Finally, $Ug_{\theta}U^Tx$ is utilized to convert $x$ back into the spatial domain.
	
	It was introduced in~\cite{defferrard2016convolutional} that $g_{\theta}$ is a function of $\Lambda$ and can be approximated as a $K$-order polynomial filter:
	
	\begin{equation}\label{polynomial filter}
	g_{\theta^\prime}(\Lambda) \approx \sum\limits_{k = 0}^{K} {\theta_k^\prime \Lambda^k}\,, 
	\end{equation}
	where $\theta_k^\prime \in \mathbb{R}^K$ is a vector of polynomial coefficients. However, computing Eq.~\ref{polynomial filter} might be prohibitively expensive for enormous number of graphs. To solve this problem, the filter is reformulated in terms of Chebyshev polynomials up to $K^{th}$ order~\cite{hammond2011wavelets}. Specifically, the Chebyshev polynomials can be calculated by a recurrence function: $T_k(a) = 2aT_{k-1}(a) - T_{k-2}(a)$ with $T_0 = 1$ and $T_1 = a$. We now have the truncated expansion of the approximated filter:
	
	\begin{equation}
	g_{\theta^{''}}(\Lambda) \approx \sum\limits_{k = 0}^{K}{\theta_k^{''}}T_k(\tilde{\Lambda})\,,
	\end{equation}
	where $\theta_k^{''} \in \mathbb{R}^K$ is a vector of Chebyshev coefficients following a scaled diagonal matrix $\tilde \Lambda = 2\Lambda/\lambda_{max} - I_n$. $\lambda_{max}$ denotes the spectral radius of $L^{sym}$. It's easy to check that $(U\Lambda U^T)^k = U\Lambda^k U^T$. Specifically, the convolution in spectral domain of vertices $x$ with a filter $g_{\theta^{''}}$ can be an equation of $\tilde{L}$:
	
	\begin{equation}\label{definition of graph convolution}
	g_{\theta^{''}} \star x \approx \sum\limits_{k = 0}^{K}{\theta_k^{''}}T_k(\tilde{L})x\,, 
	\end{equation}
	with $\tilde{L} = 2L^{sym}/\lambda_{max} -I_n$, a similar scaled strategy of $\tilde{\Lambda}$. To circumvent the problem of overfitting and build a deeper network model,~\cite{kipf2016semi} approximate $\lambda_{max} \approx 2$ and limit $K = 1$. By this means, Eq.~\ref{definition of graph convolution} further simplifies to:
	
	\begin{equation}\label{semi gcn}
	\begin{aligned}
	g_{\theta^{''}} \star x &\approx  \theta^{''}_0 x + \theta^{''}_1 (L^{sym} - I_n)x\,,
	\end{aligned}
	\end{equation}
	where $\theta^{''}_0$ and $\theta^{''}_1$ serve as two free parameters. Inspired by recent work~\cite{perozzi2014deepwalk} that random walks can be effective for learning explicit representations of vertices by preserving the structure of graph topology. The transition probability of a random walker on the graph is defined as $P = D^{-1} R \in \mathbb{R}^{n \times n}$. We normalize $P$ to obtain random walk normalization $L^{rw} \in \mathbb{R}^{n \times n}$ of graph Laplacian matrix for preventing gradient explosion:
	
	\begin{equation}\label{rw nor}
	L^{rw} = I_n - P = I_n - D^{-1}R = D^{-1}L\,,
	\end{equation}
	so that $0 \leqslant \Lambda_{ii}^{rw} \leqslant2$, where $\Lambda^{rw} \in \mathbb{R}^{n \times n}$ is a diagonal matrix which comprises eigenvalues of $L^{rw}$. Notice that:
	
	\begin{equation}
	L^{rw} = D^{-1/2}L^{sym}D^{1/2}\,,
	\end{equation}
	where $D^{1/2}$ is an invertible square matrix. It's obvious that $L^{rw}$ is similar to $L^{sym}$. For this reason, $L^{rw}$ has real eigenvalues $\Lambda_{ii}^{rw}$, even if $L^{rw}$ is in general not symmetric. Particularly, eigenvalues $\Lambda_{ii}^{rw}$ are equivalent to eigenvalues $\Lambda_{ii}$ of $L^{sym}$ and thus the approximation of Eq.~\ref{polynomial filter} is suitable for random walk normalization. Substituting $L^{rw}$ for $L^{sym}$ in Eq.~\ref{semi gcn}, we can obtain the expression of spectral graph convolutions involved with the random walk rule:
	
	\begin{equation}\label{pre-residual}
	\begin{aligned}
	g_{\theta^{''}} \star x &\approx  \theta^{''}_0 x + \theta^{''}_1 (L^{rw} - I_n)x \\
	&= \theta^{''}_0 x - \theta^{''}_1 D^{-1}Rx\,.
	\end{aligned}
	\end{equation}
	
	Instead of minimizing the number of parameters by simply hypothesizing $\theta^{''}_0 = -\theta^{''}_1$ in~\cite{kipf2016semi}, we consider Eq.~\ref{pre-residual} as a potential residual architecture comparable with ResNet~\cite{he2016deep}. Concretely, $x$ is added as a shortcut connection directly without multiplying $\theta^{''}_0$. Neglecting the negative sign in $-\theta^{''}_1$, we rewrite Eq.~\ref{pre-residual} as:
	\begin{equation}
	g_{\theta^{''}} \star x \approx \theta^{''}_1 D^{-1}Rx + x\,.
	\end{equation}
	
	Finally, we extend the spectral graph convolutions to multi-input channels with $V \in \mathbb{R}^{n\times d}$ as follows:
	
	\begin{equation}
	Z = D^{-1}RV\Theta + V\,,
	\end{equation}
	where $\Theta \in \mathbb{R}^{d\times d}$ is the filter parameter matrix. $\Theta$ can be reorganized as few stacked layers and the final output of spectral graph convolution is defined as:
	
	\begin{equation}\label{spectral graph convolution output}
	Z = (D^{-1}RVW_g)W_r + V\,,
	\end{equation}
	where $W_g \in \mathbb{R}^{d \times d}$ is the GCN weight layer and $W_r \in \mathbb{R}^{d \times d}$ is the residual weight layer. The output $Z = \{z_1,...,z_n\}, z_i \in \mathbb{R}^d$ is the relation enhanced representations for vertices. $Z$ is further processed by mean pooling to form frame-level feature $\bar{Z} \in \mathbb{R}^d$.
	
	\begin{table*}[t]
		\begin{center}
			\resizebox{\textwidth}{19mm}{
				\renewcommand{\arraystretch}{1.3} 
				\begin{tabular}{lcccccccccccc}
					\toprule
					\multirow{2}*{\textbf{Methods}} & \multicolumn{5}{c}{\textbf{Text-to-Video Retrieval}} & & \multicolumn{5}{c}{\textbf{Video-to-Text Retrieval}} & \multirow{2}*{\textbf{Sum of Recalls}}\\
					\cline{2-6}\cline{8-12}
					~ & R@1 & R@5 & R@10 & Med R & Mean R & & R@1 & R@5 & R@10 & Med R & Mean R \\
					\hline
					{VSE~\cite{kiros2014unifying}} & 5.0 & 16.4 & 24.6 & 47 & 215.1 & & 7.7 & 20.3 & 31.2 & 28 & 185.8 & 105.2 \\
					VSE++~\cite{faghri2017vse} & 5.7 & 17.1 & 24.8 & 65 & 300.8 & & 10.2 & 25.4 & 35.1 & 25 & 228.1 & 118.3 \\
					JEMC~\cite{mithun2018learning} & 5.8 & 17.6 & 25.2 & 61 & 296.6 & & 10.5 & 26.7 & 35.9 & 25 & 266.6 & 121.7 \\
					W2VV~\cite{dong2018predicting} & 6.1 & 18.7 & 27.5 & 45 & - & & 11.8 & 28.9 & 39.1 & 21 & - & 132.1 \\
					Dual encoding~\cite{dong2019dual} & 7.7 & 22.0 & 31.8 & 32 & - & & 13.0 & \textbf{30.8} & 43.3 & \textbf{15} & - & 148.6 \\
					{ViSERN} & \textbf{7.9} & \textbf{23.0} & \textbf{32.6} & \textbf{30} & \textbf{178.7} & & \textbf{13.1} & 30.1 & \textbf{43.5} & \textbf{15} & \textbf{119.1} & \textbf{151.1} \\
					\bottomrule
				\end{tabular}
			}
		\end{center}
		\caption{Video-to-Text and Text-to-Video retrieval results on the MSR-VTT dataset.} \label{results on MSR-VTT}
	\end{table*}
	
	\begin{table*}[t]
		\begin{center}
			\resizebox{\textwidth}{17mm}{
				\renewcommand{\arraystretch}{1.3}  
				\begin{tabular}{lcccccccccccc}
					\toprule
					\multirow{2}*{\textbf{Methods}} & \multicolumn{5}{c}{\textbf{Text-to-Video Retrieval}} & & \multicolumn{5}{c}{\textbf{Video-to-Text Retrieval}} & \multirow{2}*{\textbf{Sum of Recalls}}\\
					\cline{2-6}\cline{8-12}
					~ & R@1 & R@5 & R@10 & Med R & Mean R & & R@1 & R@5 & R@10 & Med R & Mean R \\
					\hline
					VSE~\cite{kiros2014unifying} & 12.3 & 30.1 & 42.3 & 14 & 57.7 & & 15.8 & 30.2 & 41.4 & 12 & 84.8 & 171.8 \\
					VSE++~\cite{faghri2017vse} & 15.4 & 39.6 & 53.0 & 9 & 43.8 & & 21.2 & 43.4 & 52.2 & 9 & 79.2 & 224.8 \\
					JEMC~\cite{mithun2018learning} & 16.1 & 41.1 & 53.5 & 9 & 42.7 & & 23.4 & 45.4 & 53.0 & 8 & 75.9 & 232.5 \\
					Dual encoding~\cite{dong2019dual} & 17.6 & 47.1 & 59.5 & 7 & 34.8 & & 20.6 & 42.8 & 58.8 & 8 & 38.9 & 244.0 \\
					{ViSERN} & \textbf{18.1} & \textbf{48.4} & \textbf{61.3} & \textbf{6} & \textbf{28.6} & & \textbf{24.3} & \textbf{46.2} & \textbf{59.5} & \textbf{7} & \textbf{34.6} & \textbf{257.8} \\
					\bottomrule
				\end{tabular}
			}
		\end{center}
		\caption{Video-to-Text and Text-to-Video retrieval results on the MSVD dataset.} \label{results on MSVD}
	\end{table*}
	
	\begin{figure*}[t]
		\centering
		\includegraphics[width=1.0\linewidth]{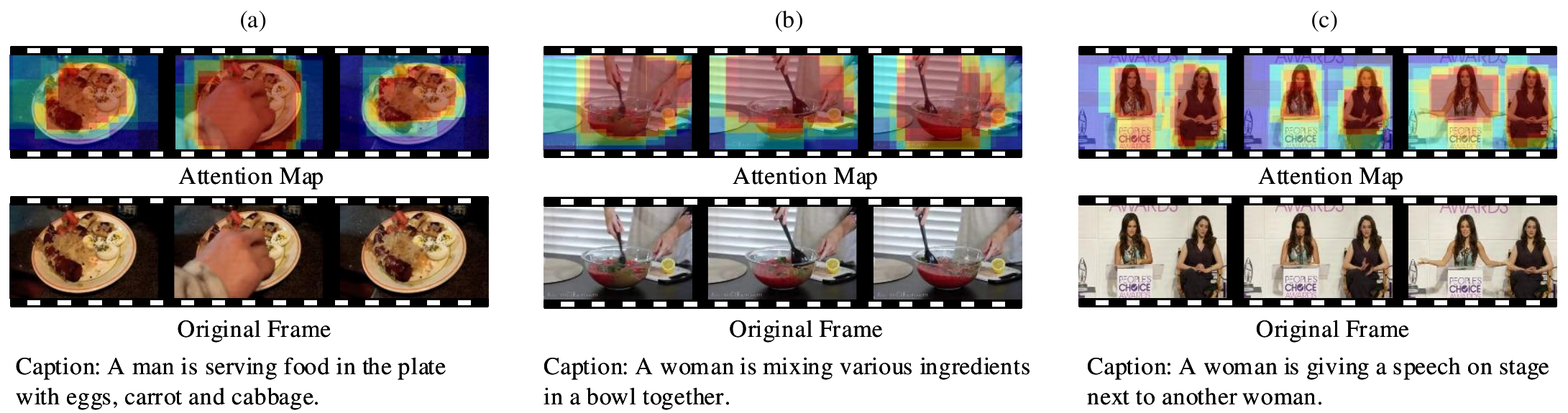}
		\caption{Visualization of attention maps and original frames above corresponding captions.}
		\label{fig:attentionframecaption}
	\end{figure*}

	\begin{table*}[t]
		\begin{center}
			\resizebox{\textwidth}{16mm}{
				\renewcommand{\arraystretch}{1.3} 
				\begin{tabular}{lcccccccccccc}
					\toprule
					\multirow{2}*{\textbf{Methods}} & \multicolumn{5}{c}{\textbf{Text-to-Video Retrieval}} & & \multicolumn{5}{c}{\textbf{Video-to-Text Retrieval}} & \multirow{2}*{\textbf{Sum of Recalls}}\\
					\cline{2-6}\cline{8-12}
					~ & R@1 & R@5 & R@10 & Med R & Mean R & & R@1 & R@5 & R@10 & Med R & Mean R \\
					\hline
					$\text{ViSERN}_{\text{no-rea}}$ & 7.5 & 22.6 & 32.4 & \textbf{30} & 187.2 & & 12.5 & \textbf{30.8} & 41.8 & 16 & 121.0 & 147.6 \\
					$\text{ViSERN}_{\text{row-nor}}$ & 7.1 & 22.1 & 31.8 & {32} & 193.4 & & 12.1 & 30.6 & 42.2 & 16 & 128.6 & 145.9 \\
					$\text{ViSERN}_{\text{sym-nor}}$ & 7.6 & 22.8 & 32.3 & \textbf{30} & 189.5 & & 12.1 & 30.7 & 42.6 & 16 & 119.6 & 148.0 \\
					ViSERN  & \textbf{7.9} & \textbf{23.0} & \textbf{32.6} & \textbf{30} & \textbf{178.7} & & \textbf{13.1} & 30.1 & \textbf{43.5} & \textbf{15} & \textbf{119.1} & \textbf{151.1}  \\
					\bottomrule
				\end{tabular}
			}
		\end{center}
		\caption{Ablation studies on the MSR-VTT dataset.} 
		\label{ablation study}
	\end{table*}
	
	\subsection{Common Space Learning}\label{Common Space Learning}
	For both video and text ultimate embedding representations, we use the multi-level encoder~\cite{dong2019dual} to map them in the same $D$-dimensional common space. We denote the affined video and text features as $O$ and $T$, respectively.
	
	We adopt a hinge-based triplet ranking loss~\cite{faghri2017vse} which is proved to be powerful for common space learning with concentration on hard negatives. The loss is expressed as:
	
	\begin{equation}\label{triplet loss}
	\begin{aligned}
	L(O, T) &= [\alpha-S(O, T) + S(\hat{O}, T)]_{+} \\
	&+ [\alpha-S(O, T) + S(O, \hat{T})]_{+}\,,
	\end{aligned}
	\end{equation}
	where $\alpha$ is a margin constant, while $S(\cdot)$ computes cosine similarity between two features and $[x]_{+} \equiv \max(x, 0)$. Given a positive pair $(O, T)$ within a mini-batch, $\hat{O} = {\arg \max}_{O \ne i}S(i, T)$ and $\hat{T} = {\arg \max}_{T \ne j}S(O, j)$ are the hardest negative samples.
	
	\section{Experiment}
	To demonstrate the effectiveness of the proposed ViSERN model, we conduct extensive experiments on the video-text retrieval task (i.e., video query and text query as input respectively to search the counterparts). Two publicly available datasets are performed in our experiments. We analyze the results comparing with recent state-of-the-art methods. Moreover, we estimate the performance of each component in our model by ablation studies.
	
	\subsection{Datasets and Evaluation Metric}
	We evaluate our model on two benchmark datasets: the MSR-VTT dataset~\cite{xu2016msr} and the MSVD dataset~\cite{chen2011collecting}. The MSR-VTT dataset is composed of 10k video clips with 20 sentence captions per clip, one of the largest datasets in terms of video-text pairs and word vocabulary. The MSVD dataset consists of 1970 video clips with around 40 multilingual sentence captions per clip. We only consider the English captions and stochastically choose five captions from it. Both two datasets are collected from Youtube in the range of multiple categories. We follow the identical partition strategy in~\cite{dong2019dual} for training, testing and validation in our experiments.
	
	The evaluation criteria in our experiments are the rank-based performance generally listed as Recall at K ({R@K}, K = 1, 5, 10), Median Rank ({Med R}) and Mean Rank ({Mean R}). Given the retrieval results, R@K is the percentage of queries whether at least one corresponding item shows in the top-K. Med R calculates the median rank of the first correct item among the search results. Similarly, Mean R computes the mean rank of whole correct results. The sum of R@1, R@5 and R@10, noted as Sum of Recalls is also reported.
	
	\subsection{Implementation Details}
	We uniformly sample 16 frames from videos with the same time interval between every two frames. The number $n$ of regions within a frame is 36, identical to~\cite{anderson2018bottom}. The dimension $d$ of region features extracted from ResNet-101 is 2048. The detailed configuration of the bottom-up attention model is the same as~\cite{anderson2018bottom}. For text feature initial representations, each word is firstly represented by a one-hot vector and then multiplied by a word embedding matrix. We set the word embedding size to 500 and the dimension of the common space $D$ to 2048, similar to~\cite{dong2019dual}. The margin parameter $\alpha$ is empirically chosen to be 0.2. The size of a mini-batch is 64. The optimizer in the training procedure is Adam with 50 epochs at most. We start training with an initial learning rate 0.0001, and the adjustment schedule is that once the validation loss does not decrease in three consecutive epochs, the learning rate is divided by 2.
	
	\subsection{Comparisons with the State-of-the-art}
	Quantitative results on the MSR-VTT and the MSVD datasets are shown in Table~\ref{results on MSR-VTT} and Table~\ref{results on MSVD}, respectively. We can observe that the proposed ViSERN outperforms recent state-of-the-art methods in terms of most indicators on both datasets, especially for text-to-video retrieval subtask, which has more actual applications in daily life. For example, our ViSERN improves 2.6\%, 4.5\% and 2.5\% at R@1 and R@5 and R@10 respectively on the MSR-VTT dataset comparing to the current state-of-the-art. In addition, we notice that ViSERN obtains much gain on the MSVD dataset relatively. The reasons are probably considered as two folds: First, the size of the MSVD dataset is smaller than the MSR-VTT dataset. Indeed, it potentially results from the existence of fewer distractors for a given query in the MSVD dataset. Second, the MSVD dataset only provides the visual signal, while the MSR-VTT dataset offers both visual and auditory information. It is non-trivial to exploit comprehensive elements in the MSR-VTT dataset for better representations. The results demonstrate that our ViSERN model can usefully implement semantic visual reasoning and measure the video-text similarity more accurately.
	
	\subsection{Ablation Studies}
	To validate the contribution of each component in our model, we carry out several ablation experiments. We start with a fundamental option which is simply implemented without the random walk rule-based GCN as a baseline model (denoted as $\text{ViSERN}_{\text{no-rea}}$). The region features $V$ are straightforwardly processed by mean pooling to represent the whole frame without performing reasoning. To exam the benefit of reasoning based on graph topology, we use row-wise normalization  (denoted as $\text{ViSERN}_{\text{row-nor}}$) in Eq.~\ref{spectral graph convolution output}, a similar method avoiding gradient explosion implemented in~\cite{li2019visual}. Furthermore, we confirm the effectiveness of random walk statistics by comparing with symmetric normalization (denoted as $\text{ViSERN}_{\text{sym-nor}}$).
	
	The detailed ablation results on the MSR-VTT dataset are shown in Table~\ref{ablation study}. From the results of $\text{ViSERN}_{\text{no-rea}}$, we observe that it is reasonable to represent the whole frame by 36 region features. The results of $\text{ViSERN}_{\text{row-nor}}$ indicate that reasoning is substantially dependent on graph topology-based normalization. We also notice that random walk statistics help to improve the performance by comparing the results of $\text{ViSERN}_{\text{sym-nor}}$ with ViSERN. One possible reason could be that the random walk rule is effective to maintain the structure of graph topology generated from each individual frame and maximize the contribution of shared GCN layer weights and residual layer weights.
	
	\subsection{Visualization and Analysis}
	The motivation of our visual semantic reasoning is to highlight the regions with strongly semantic information by exploiting relations within a frame. In order to validate our thoughts, we visualize the correlations between the feature of the whole frame and the region features included in this frame by applying an attention map. Specifically, we compute the cosine similarity scores (same as $S(\cdot)$ in Eq.~\ref{triplet loss}) between each region feature $z_i$ and the feature of entire frame $\bar{Z}$. Then we rank the regions in the descending order according to similarity scores. To generate the attention map, we assign an attention score $s_i \in \mathbb{R}$ to each region based on its rank $r_i \in \{0, 1,..., n-1\}$. The function of attention score is defined as $s_i = (n - r_i)^2$. We normalize the set of attention scores to generate the final attention map conveniently.
	
	In Figure~\ref{fig:attentionframecaption}, we show the attention maps, original frames and corresponding captions. The regions with strongly semantic meaning are painted red in attention maps. From the attention visualization, we can observe that ViSERN well represents frame-level features by exploiting semantic reasoning and capturing regions with intensely semantic relations. Additionally, Figure~\ref{fig:attentionframecaption}(a) indicates that ViSERN is sensitive to short-term interaction. We also find that long-term motion is also captured from the illustration of Figure~\ref{fig:attentionframecaption}(b). Furthermore, the regions of both two salient people are highlighted separately in Figure~\ref{fig:attentionframecaption}(c).
	
	\section{Conclusion}
	In this paper, we propose a Visual Semantic Enhanced Reasoning Network (ViSERN) to learn semantic reasoning by using the novel random walk rule-based graph convolutional networks for video-text retrieval. The enhanced frame representations capture key regions and suppress redundancy in a scene so that it is helpful to align with the corresponding video caption. Extensive experiments on the MSR-VTT and MSVD datasets demonstrate the proposed model consistently outperforms the state-of-the-art methods. In addition, we provide visualization analyses to show how ViSERN can provide more discrimination and interpretability to such vision-language models. Moreover, ViSERN can be further developed to support content-based searches on online video websites. 
	
	In the future, we will consider the location information of regions representing in the graph. We would also like to extend the undirected graph to a directed graph for modeling video temporal sequences reasonably.
	
	\section*{Acknowledgments}
	This work was partially supported by the China Telecom Dict Application Capability Center, the Beijing Natural Science Foundation (4202049) and the National Key R\&D Program of China (2018YFB1800805).
	
	\newpage

	\bibliographystyle{named}
	\bibliography{ViSERN}
	
\end{document}